\documentclass[runningheads,DRAFT]{llncs}

\usepackage{graphicx}
\usepackage{amsmath,amssymb,amsfonts,nicefrac,derivative}

\newcommand{\nR}{\mathbb{R}}
\newcommand{\dif}[1]{\mathrm{d}{#1}\,}
\newcommand{\CK}{\text{CK}}
\newcommand{\strain}{\text{strain}}

\usepackage{csquotes}
\usepackage{cleveref}
\usepackage{caption}
\usepackage{subcaption}
\usepackage{enumitem}

\usepackage{listings}
\usepackage{tikz}
\tikzset{>=latex}

\usepackage{epstopdf}

\usepackage{url}

\usepackage[acronym]{glossaries}
\usepackage{glossaries-extra}
\setabbreviationstyle[acronym]{long-short}
\newacronym{ml}{ML}{Machine Learning}

\begin{document}
\title{
  $\mathcal{C}^k$-continuous Spline Approximation with TensorFlow Gradient
  Descent Optimizers%
  \thanks{Stefan Huber and Hannes Waclawek are supported by the European
  Interreg Österreich-Bayern project AB292 KI-Net and the Christian Doppler
  Research Association. This preprint has not undergone peer review or any
  post-submission improvements or corrections. The Version of Record of this
  contribution is published in Computer Aided Systems Theory – EUROCAST 2022 and
  is available online at
  \protect\url{https://doi.org/10.1007/978-3-031-25312-6_68}.}
}
\titlerunning{Spline Approximation with TensorFlow Gradient Descent Optimizers}

%
%
\author{Stefan Huber \and Hannes Waclawek}
\authorrunning{S. Huber \and H. Waclawek}
%
\institute{Salzburg University of Applied Sciences, Salzburg, Austria \\
	\email{\{stefan.huber, hannes.waclawek\}@fh-salzburg.ac.at}}
\maketitle              

\begin{abstract}
  In this work we present an \enquote{out-of-the-box} application of Machine
  Learning (ML) optimizers for an industrial optimization problem. We introduce
  a piecewise polynomial model (spline) for fitting of $\mathcal{C}^k$-continuos
  functions, which can be deployed in a cam approximation setting. We then use
  the gradient descent optimization context provided by the machine learning
  framework TensorFlow to optimize the model parameters with respect to
  approximation quality and $\mathcal{C}^k$-continuity and evaluate available
  optimizers. Our experiments show that the problem solution is feasible using
  TensorFlow gradient tapes and that AMSGrad and SGD show the best results among
  available TensorFlow optimizers. Furthermore, we introduce a novel
  regularization approach to improve SGD convergence. Although experiments show
  that remaining discontinuities after optimization are small, we can eliminate
  these errors using a presented algorithm which has impact only on affected derivatives
  in the local spline segment.
  \keywords{Gradient Descent Optimization \and TensorFlow
  \and Polynomial Approximation \and Splines \and Regression}
\end{abstract}

\section{Introduction}

When discussing the potential application of \gls{ml} to industrial settings,
we first of all have the application of various \gls{ml} methods and models
per se in mind.
These methods, from neural networks to simple linear classifiers, are based on
gradient descent optimization. This is why \gls{ml} frameworks come with a variety
of gradient descent optimizers that perform well on a diverse set of problems
and in the past decades have received significant improvements in academia and
practice.

Industry is full of classical numerical optimization and we can
therefore, instead of using the entire framework in an industrial context,
harness modern optimizers that lie at the heart of modern ML methods directly and
apply them to industrial numerical optimization tasks. One of these optimization
tasks is cam approximation, which is the task of fitting a continuous function
to a number of input points with properties favorable for cam design. One way to
achieve these favorable properties is via gradient based approaches, where an
objective function allows to minimize user-definable losses. Servo drives like
B\&R Industrial Automation's ACOPOS series process cam profiles as a piecewise
polynomial function (spline). This is why, with the goal of using the findings
of this paper as a basis for cam approximation in future works, we want to lay
the ground for performing polynomial approximation with a
$\mathcal{C}^k$-continuous piecewise polynomial spline model using gradient
descent optimization provided by the machine learning framework TensorFlow. The
continuity class $\mathcal{C}^k$ denotes the set of $k$-times continuously
differentiable functions $\nR \to \nR$. Continuity is important in cam design
concerning forces that are only constrained by the mechanical construction of
machine parts. This leads to excessive wear and vibrations which we ought to
prevent. Although our approach is motivated by cam design, it is generically
applicable.

The contribution of this work is manifold:

\begin{enumerate}
  \item \enquote{Out-of-the-box} application of ML-optimizers for an industrial
  setting.
\item A $\mathcal{C}^k$-spline approximation method with novel gradient
  regularization.
  \item Evaluation of TensorFlow optimizer performance for a well-known problem.
  \item Non-convergence of optimizers using exponential moving averages, like
  Adam, is documented in literature \cite{reddi2019}. We confirm with our
  experiments that this non-convergence extends to the presented optimization
  setting.
  \item Algorithm to strictly establish continuity with impact only on affected
  derivatives in the local spline segment.
\end{enumerate}

The Python libraries and Jupyter Notebooks used to perform our
experiments are available under an MIT license at \cite{waclawek2022}.

\paragraph{Prior work}
There is a lot of prior work on neural networks for function approximation
\cite{adcock2021} or the use of gradient descent optimizers for B-spline curves
\cite{sandgren1989}. There are also non-scientific texts on gradient descent
optimization for polynomial regression. However, to the best of our knowledge,
there is no thorough evaluation of gradient descent optimizers for
$\mathcal{C}^k$-continuous piecewise polynomial spline approximation.

\section{Gradient descent optimization}

TensorFlow provides a mechanism for automatic gradient computation using a
so-called gradient tape. This mechanism allows to directly make use of the
diverse range of gradient based optimizers offered by the framework and
implement custom training loops. We implemented our own training loop in which
we (i) obtain the gradients for a loss expression $\ell$, (ii) optionally apply
some regularization on the gradients and (iii) supply the optimizer with the
gradients.
This requires a computation of $\ell$ that allows the gradient tape to track
the operations applied to the model parameters in form of TensorFlow variables.

\subsection{Spline model} \label{sec:spline_model}

\begin{figure}[tb]
  \centering
  \includegraphics[width=\textwidth, page=2]{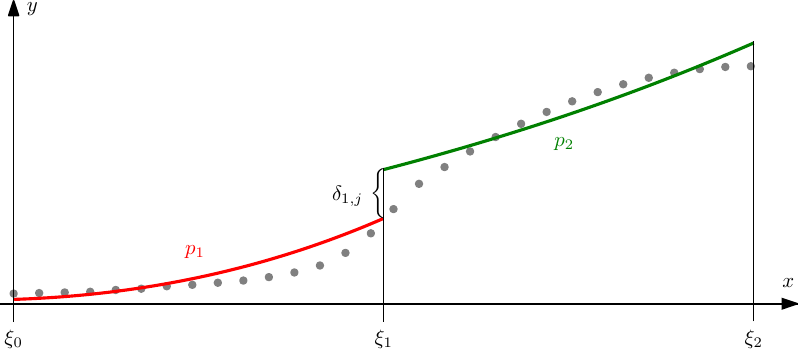}
  \caption{Spline model derivative $j$ consisting of two polynomial segments.}
  \label{fig:spline_model}
\end{figure}

Many servo drives used in industrial applications, like B\&R Industrial
Automation's ACOPOS series, use piecewise polynomial functions
(splines) as a base model for cam follower displacement curves. This requires
the introduction of an according spline model in TensorFlow. Let us consider $n$
samples at $x_1 \le \dots \le x_n \in \nR$ with respective values $y_i \in \nR$.
We ask for a spline $f \colon I \to \nR$ on the interval $I = [x_1, x_n]$ that
approximates the samples well and fulfills some additional domain-specific
properties, like $\mathcal{C}^k$-continuity or
$\mathcal{C}^k$-cyclicity\footnote{By $\mathcal{C}^k$-cyclicity we mean that the
derivative $f^{(i)}$ matches on $x_1$ and $x_n$ for $1 \le i \le k$. If it
additionally matches for $i=0$ then we have $\mathcal{C}^k$-periodicity.}.
Let us denote by $\xi_0 \le \dots \le \xi_m$ the polynomial boundaries of $f$,
where $\xi_0 = x_1$ and $\xi_m = x_n$. With $I_i = [\xi_{i-1}, \xi_i]$, the
spline $f$ is modeled by $m$ polynomials $p_i \colon I \to \nR$ that agree with
$f$ on $I_i$ for $1 \le i \le m$. 
Each polynomial 
\begin{align}
  \label{eq:polynomial_model}
  p_i = \sum_{j=0}^d \alpha_{i,j} x^j 
\end{align}
is determined by its coefficients $\alpha_{i,j}$, where $d$ denotes the degree
of the spline and its polynomials. That is, the $\alpha_{i,j}$ are the to be
trained model parameters of the spline ML model. We investigate the convergence
of these model parameters $\alpha_{i,j}$ of this spline model with respect to
different loss functions, specifically for $L_2$-approximation error and
$\mathcal{C}^k$-continuity, by means of different TensorFlow optimizers (see
details below). Figure \ref{fig:spline_model} depicts the principles of the
presented spline model.

\subsection{Loss Function} \label{sec:loss_function}

In order to establish $\mathcal{C}^k$-continuity, cyclicity, periodicity and
allow for curve fitting via least squares approximation, we introduce the cost
function

\begin{align}
  \label{eq:total_loss}
    \ell = \lambda \ell_2 + (1-\lambda) \ell_\CK.
\end{align}

By adjusting the value of $\lambda$ in equation \eqref{eq:total_loss} we can put
more weight to either the approximation quality or $\mathcal{C}^k$-continuity
optimization target.
%
The approximation error $\ell_2$ is the least-square error and made invariant
to the number of data points and number of polynomial segments by the following
definition:

\begin{align}
  \label{eq:l2_loss}
  \ell_2 = \frac{m}{n} \sum_i |f(x_i) - y_i|^2.
\end{align}

We assign a value $\ell_\CK$ to the amount of discontinuity in our spline
function by summing up discontinuities at all $\xi_i$ across relevant
derivatives as

\begin{align}
  \label{eq:ck_loss}
  \ell_\CK = \frac{1}{m - 1} \sum_{i=1}^{m-1} \sum_{j=0}^k \delta_{i,j}^2
  \quad \text{with} \quad
   \delta_{i,j} = p^{(j)}_{i+1}(\xi_i) - p^{(j)}_i(\xi_i).
\end{align}



We make $\ell_\CK$ in equation \eqref{eq:ck_loss} invariant to the
number of polynomial segments by applying an equilibration factor $\frac{1}{m -
1}$, where $m - 1$ is the number of boundary points excluding $\xi_0$ and
$\xi_m$.
%
%
This loss $\ell_\CK$ can be naturally extended to
$\mathcal{C}^k$-cyclicity/periodicity for cam profiles.\footnote{In
\eqref{eq:ck_loss}, change $m - 1$ to $m$ and generalize $\delta_{i,j} =
p^{(j)}_{1 + (i \bmod m)}(\xi_{i \bmod m}) - p^{(j)}_i(\xi_i)$. For cyclicity
we ignore the case $j=0$ when $i=m$, but not for periodicity.}

\subsection{TensorFlow Training Loop}

The gradient tape environment of TensorFlow offers automatic differentiation of
our loss function defined in equation \eqref{eq:total_loss}. This requires a
computation of $\ell$ that allows for tracking the operations applied to
$\alpha_{i,j}$ through the usage of TensorFlow variables and arithmetic
operations, see Listing \ref{lst:optimization_loop}.

In this training loop, we first calculate the loss according to equation
\eqref{eq:total_loss} in a gradient tape context in lines
\ref{lst:optimization_loop_loss1} and 
\ref{lst:optimization_loop_loss2}
and then obtain the gradients according to that loss result in line
\ref{lst:optimization_loop_gradient} via the gradient tape environment
automatic differentiation mechanism. We then apply regularization in line
\ref{lst:optimization_loop_regularization} that later will be introduced in
section \ref{sec:degree_based_regularization} and supply the optimizer with the
gradients in line \ref{lst:optimization_loop_apply_grads}.

\newpage

\begin{lstlisting}[caption={Gradient descent optimization loop in TensorFlow.}, 
                  label=lst:optimization_loop]
  for e in range(epochs):
      with tf.GradientTape(persistent=True) as tape:
          loss_l2 = calculate_l2_loss()
          loss_ck = calculate_ck_loss()
          loss = tf.add(tf.multiply(loss_l2, lambd), |\label{lst:optimization_loop_loss1}|
                        tf.multiply(loss_ck, 1.0-lambd))|\label{lst:optimization_loop_loss2}|
      gradients = tape.gradient(loss, coeffs) |\label{lst:optimization_loop_gradient}|
      gradients = apply_regularization() |\label{lst:optimization_loop_regularization}|
      optimizer.apply_gradients(zip(gradients, coeffs)) |\label{lst:optimization_loop_apply_grads}|
\end{lstlisting}

\section{Improving spline model performance} \label{sec:improve_spline_model}

In order to improve convergence behavior using the model defined in section
\ref{sec:spline_model}, we introduce a novel regularization approach and
investigate effects of input data scaling and shifting of polynomial centers. In
a cam design context, discontinuities remaining after the optimization procedure
lead to forces and vibrations that are only constrained by the cam-follower
system's mechanical design. To prevent such discontinuities, we propose an
algorithm to strictly establish continuity after optimization.

\subsection{A degree-based regularization} \label{sec:degree_based_regularization}

With the polynomial model described in equation \eqref{eq:polynomial_model},
terms of higher order have greater impact on the result. This leads to gradients
having greater impact on terms of higher order, which impairs convergence
behavior. This effect is also confirmed by our experiments. We propose a
degree-based regularization approach, that mitigates this impact by effectively
causing a shift of optimization of higher-degree coefficients to later
epochs. We do this by introducing a gradient regularization vector $R = (r_0,
\dots, r_d)$, where

\begin{align}
  r_j = \frac{r'_j}{\sum_{k=0}^d r'_k}
  \quad \text{with} \quad
  r'_j = \frac{1}{1 + j}.
\end{align}
%
The regularization is then applied by multiplying each gradient value
$\pdv{\ell}{\alpha_{i,j}}$ with $r_j$. Since the entries $r_j$ of $R$ sum up to
$1$, this effectively acts as an equilibration of all gradients per polynomial
$p_i$.

This approach effectively makes the sum of gradients degree-independent.
Experiments show that this allows for higher learning rates using non-adaptive
optimizers like SGD and enables the use of SGD with Nesterov momentum, which
does not converge without our proposed regularization approach. This brings
faster convergence rates and lower remaining losses for non-adaptive
optimizers. At a higher number of epochs, the advantage of the regularization
is becoming less. Also, the advantage of the regularization is higher for
polynomials of higher degree, say, $d \geq 4$.

\subsection{Practical considerations} \label{sec:practical_considerations}

Experiments show that, using the training parameters outlined in chapter
\ref{chap:experiments}, SGD optimization has a certain radius of convergence
with respect to the $x$-axis around the polynomial center. 
Shifting of polynomial centers to the mean of the respective segment allows segments with
higher $x$-value ranges to converge. We can implement this by extending the
polynomial model defined in equation \eqref{eq:polynomial_model} as

\begin{align}
  \label{eq:polynomial_model_shifting_center}
  p_i = \sum_{j=0}^d \alpha_{i,j} (x-\mu_i)^j,
  \quad \text{where} \quad
  \mu_i = \frac{\xi_{i-1}+\xi_{i}}{2}.
\end{align}


If input data is scaled such that every polynomial segment is in the range $[0,
1]$, in all our experiments for all $0 \leq \lambda \leq 1$, SGD optimization is
able to converge using this approach. With regards to scaling, as an example,
for a spline consisting of $8$ polynomial segments, we scale the input data such
that $I = [0, 8]$. We skip the back-transformation as we would do in production
code.

\subsection{Strictly establishing continuity after optimization} \label{sec:continuity_algorithm}

In order to strictly establish $\mathcal{C}^k$-continuity after optimization,
i.e., to eliminate possible remaining $\ell_\CK$, we apply corrective
polynomials that enforce $\delta(\xi_i) = 0$ at all $\xi_i$.
The following method requires a spline degree $d \ge 2k + 1$. Let

\begin{align}
  \label{eq:mean_derivative}
  m_{i,j} = \frac{p^{(j)}_i(\xi_i) + p^{(j)}_{i+1}(\xi_i)}{2}
\end{align}

denote the mean $j$-th derivative of $p_i$ and $p_{i+1}$ at $\xi_i$ for all $0
\le j \le k$. Then there is a unique polynomial $c_i$ of degree $2k+1$ that has
a $j$-th derivative of $0$ at $\xi_{i-1}$ and $m_j - p^{(j)}_i(\xi_i)$ at
$\xi_i$ for all $0 \le j \le k$. Likewise, there is a unique polynomial
$c_{i+1}$ with $j$-th derivative given by $m_j - p^{(j)}_{i+1}(\xi_i)$ at
$\xi_i$ and $0$ at $\xi_{i+1}$. The corrected polynomials $p^*_i = p_i + c_i$
and $p^*_{i+1} = p_{i+1} + c_{i+1}$ then possess identical derivatives $m_j$ at
$\xi_i$ for all $0 \le j \le k$, yet, the derivatives at $\xi_{i-1}$ and
$\xi_{i+1}$ have not been altered. This allows us to apply the corrections at
each $\xi_i$ independently as they have only local impact. This is a nice
property in contrast to natural splines or methods using B-Splines as discussed
in \cite{sandgren1989}.

\section{Experimental Results} \label{chap:experiments}

In a first step, we investigated mean squared error loss by setting
$\lambda = 1$ in our loss function defined in equation \eqref{eq:total_loss} for
a single polynomial, which revealed a learning rate of $0.1$ as a reasonable
setting. We then ran tests with available TensorFlow optimizers listed in
\cite{tf2022} and compared their outcomes. We found that SGD with momentum,
Adam, Adamax as well as AMSgrad show the lowest losses, with a declining
tendency even after $5000$ epochs. However, the training curves of Adamax and
Adam exhibit recurring phases of instability every $\sim500$ epochs.
Non-convergence of these optimizers is documented in literature
\cite{reddi2019} and we can confirm with our experiments that it also extends
to our optimization setting. Using the AMSGrad variant of Adam eliminates this
behavior with comparable remaining loss levels. With these results in mind, we
chose SGD with Nesterov momentum as non-adaptive and AMSGrad as adaptive
optimizer for all further experiments, in order to work with optimizers from
both paradigms.

\begin{figure}[h]
  \centering
  \includegraphics[width=1\textwidth]{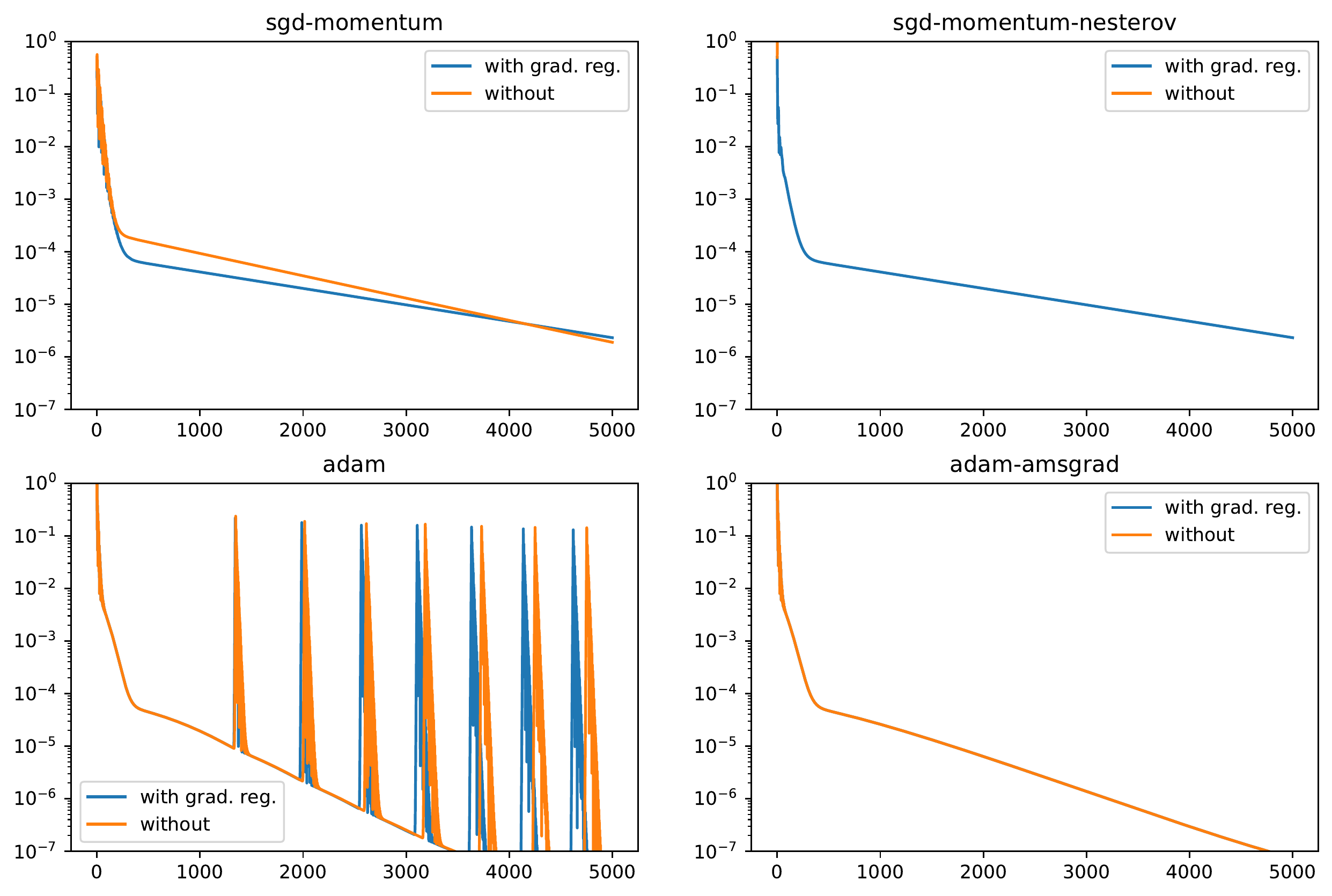}
  \caption{Overview of optimizer loss curves for one polynomial with $\lambda = 1$.}
  \label{fig:optimizers_overview}
\end{figure}

The AMSGrad optimizer performs better on the $\lambda = 1$ optimization
target, however, SGD is competitive. The loss curves of these optimizer
candidates, as well as instabilities in the Adam loss curve are shown in figure
\ref{fig:optimizers_overview}. An overview of all evaluated optimizers is given
in our GitHub repository at \cite{waclawek2022}.

With our degree-based regularization approach introduced in section
\ref{sec:degree_based_regularization}, SGD with momentum is able to converge
quicker and we are able to use Nesterov momentum, which was not possible
otherwise. We achieved best results with an SGD momentum setting of $0.95$ and
AMSGrad $\beta_1$ = 0.9, $\beta_2$=0.999 and $\epsilon=10^{-7}$.
On that basis, we investigated our general spline model, by we sweeping
$\lambda$ from $1$ to $0$. Experiments show that both optimizers are able to
generate near $\mathcal{C}^2$-continuos results across the observed
$\lambda$-range while at the same time delivering favorable approximation
results. The remaining continuity correction errors for the algorithm introduced
in section \ref{sec:continuity_algorithm} to process are small.


Using $\mathcal{C}^2$-splines of degree $5$, again, AMSGrad has a better
performance compared to SGD. For all tested $0 < \lambda < 1$, SGD and AMSGrad
manage to produce splines of low loss within 10\,000 epochs: SGD reaches $\ell
\approx 10^{-4}$ and AMSGrad reaches $\ell \approx 10^{-6}$. Given an
application-specific tolerance, we may already stop after a few hundred epochs.

\section{Conclusion and Outlook}

We have presented an \enquote{out-of-the-box} application of ML optimizers for
the industrial optimization problem of cam approximation. The model introduced
in section \ref{sec:spline_model} and extended by practical considerations in
section \ref{sec:practical_considerations} allows for fitting of
$\mathcal{C}^k$-continuos splines, which can be deployed in a cam approximation
setting. Our experiments documented in section \ref{chap:experiments} show that
the problem solution is feasible using TensorFlow gradient tapes and that
AMSGrad and SGD show the best results among available TensorFlow optimizers. Our
gradient regularization approach introduced in section
\ref{sec:degree_based_regularization} improves SGD convergence and allows usage
of SGD with Nesterov momentum. Although experiments show that remaining
discontinuities after optimization are small, we can eliminate these errors
using the algorithm introduced in section \ref{sec:continuity_algorithm}, which
has impact only on affected derivatives in the local spline segment.

Additional terms in $\ell$ can accommodate for further domain-specific goals.
For instance, we can reduce oscillations in $f$ by penalizing the strain energy

\begin{equation*}
    \ell_{\strain} = \int_I f''(x)^2 \; \dif x.
\end{equation*}

In our experiments outlined in the previous section, we started with all
polynomial coefficients initialized to zero to investigate convergence. To
improve convergence speed in future experiments, we can start with the
$\ell_2$-optimal spline and let our method minimize the overall goal $\ell$.

Flexibility of our method with regards to the underlying polynomial model allows
for usage of different function types. In this way, as an example, an orthogonal 
basis using Chebyshev polynomials could improve convergence behavior compared to 
classical monomials.

\newpage

%
%
%
\bibliographystyle{splncs04}
\bibliography{references}

\end{document}